\documentclass[conference]{IEEEtran}
\IEEEoverridecommandlockouts

\usepackage{cite}
\usepackage{url}
\usepackage{makecell}
\usepackage{amsmath,amssymb,amsfonts}
\usepackage{algorithmic}
\usepackage{graphicx}
\usepackage{textcomp}
\usepackage{xcolor}
\def\BibTeX{{\rm B\kern-.05em{\sc i\kern-.025em b}\kern-.08em
    T\kern-.1667em\lower.7ex\hbox{E}\kern-.125emX}}
\begin{document}

\title{A Domain-Structured Ensemble Framework for Perioperative Outcome Prediction Using Electronic Health Record Data\\
}

\author{\IEEEauthorblockN{Shikhar Shukla}
\IEEEauthorblockA{\textit{BioHealth Informatics} \\
\textit{Indiana University Indianapolis}\\
Indianapolis, USA \\
0009-0003-8941-3396}
\and
\IEEEauthorblockN{Cristina Barboi}
\IEEEauthorblockA{\textit{Department of Anesthesiology} \\
\textit{Indiana University School of Medicine}\\
Indianapolis, USA \\
0000-0002-8356-7285}
}

\maketitle

\begingroup
\renewcommand{\thefootnote}{}
\footnotetext{\scriptsize
© 2026 IEEE.  Personal use of this material is permitted.  Permission from IEEE must be obtained for all other uses, in any current or future media, including reprinting/republishing this material for advertising or promotional purposes, creating new collective works, for resale or redistribution to servers or lists, or reuse of any copyrighted component of this work in other works.
}
\endgroup

\begin{abstract}
Perioperative risk prediction models are often limited by narrow surgical populations, incomplete intraoperative data, poor calibration, and limited interpretability. We present a domain-structured ensemble framework for perioperative outcome prediction using routinely collected electronic health record (EHR) data, designed for extensibility across diverse clinical endpoints. The framework organizes predictors into three clinically motivated domains: patient-related (baseline vulnerability), surgery-related (procedural characteristics), and anesthetics-related (intraoperative exposures and physiologic perturbations). Domain-specific gradient boosting models generate independent risk estimates, which are integrated through a logistic regression meta-learner to produce calibrated predictions. We demonstrate the framework using postoperative delirium (POD) as an exemplar application in a case--control sample of 5,386 surgical encounters (2,693 cases, 2,693 controls) from a statewide health information exchange. POD was identified through dual confirmation requiring both ICD codes and positive Confusion Assessment Method screening within seven postoperative days; patients with preexisting dementia were excluded. The stacked meta-learner achieved an area under the receiver operating characteristic curve (AUROC) of 0.899 (95\% CI: 0.891--0.906), precision-recall AUC of 0.881, and Brier score of 0.126, compared with 0.849 for the best single-stage model. Domain ablation analysis confirmed that the three-domain architecture improved both discrimination and calibration relative to a surgery-only model (AUROC 0.879, Brier 0.140). Temporal validation on held-out post-2017 data yielded AUROC of 0.915. Calibration was excellent: intercept $-0.006$ (95\% CI: $-0.083$ to 0.070), slope 1.035 (95\% CI: 0.982 to 1.088). Decision curve analysis, corrected for case--control sampling, demonstrated positive net benefit across clinically plausible risk thresholds. The framework's modular architecture supports substitution of outcome definitions, extension of predictor domains, and dynamic risk updating as perioperative data accrue. With external validation, this approach may serve as a scalable foundation for interpretable, calibration-aware perioperative clinical decision support across multiple postoperative outcomes.
\end{abstract}


\section{Introduction}

Adverse perioperative outcomes remain a major source of morbidity, prolonged hospitalization, mortality, and healthcare cost across surgical populations~\cite{ref1,ref2}. Globally, an estimated 313 million surgical procedures are performed annually~\cite{ref3}, and recent estimates suggest that at least 4.2 million people die within 30 days of surgery each year, making postoperative death the third leading contributor to global mortality~\cite{ref4}. Perioperative organ injury affects approximately 4\% of surgical patients and is associated with a nine-fold increase in mortality risk~\cite{ref2}. Despite advances in surgical and anesthetic care, many clinically important perioperative conditions remain under-recognized or inconsistently detected in routine practice, particularly when manifestations are transient, fluctuating, or heterogeneous in presentation~\cite{ref5}. Variability in detection methods, ranging from structured clinical assessments to administrative coding, further complicates reliable outcome ascertainment~\cite{ref6}. These challenges underscore the need for robust, generalizable prediction frameworks to support early risk stratification and targeted prevention in perioperative care.

Numerous perioperative risk prediction models have been proposed; however, their translation into clinical practice has been limited by recurring methodological shortcomings~\cite{ref7,ref8}. Many models are derived from single-center cohorts or restricted to specific procedures, limiting generalizability across diverse surgical contexts~\cite{ref9}. Others rely primarily on preoperative data, underutilizing intraoperative factors such as anesthetic exposure, physiologic perturbations, surgical duration, and procedural complexity, despite growing evidence that perioperative risk evolves dynamically across the surgical course~\cite{ref10,ref11}. Traditional tools such as the American Society of Anesthesiologists (ASA) classification and the Revised Cardiac Risk Index (RCRI) offer limited discriminatory power and adaptability across outcomes~\cite{ref12,ref13}. In addition, model evaluation frequently emphasizes discrimination while neglecting calibration and clinical utility, limiting confidence in the use of predicted probabilities for real-world decision-making~\cite{ref14,ref15}.

Recent applications of machine learning have improved predictive discrimination for perioperative outcomes; however, many remain limited by poor interpretability, insufficient attention to calibration, and lack of decision-analytic evaluation~\cite{ref9,ref16}. A recent systematic review of 103 machine learning studies in perioperative medicine found that 75\% were based on single-center validations, only 13\% were externally validated, and the majority demonstrated high risk of bias~\cite{ref9}. Few models explicitly assess whether predictions meaningfully improve clinical decision-making compared with default strategies, and even fewer demonstrate prospective clinical impact~\cite{ref14,ref17}. Moreover, evaluation designs that fail to account for patient-level dependence or biased sampling can inflate performance estimates, further hindering reliable implementation~\cite{ref18}.

A domain-based modeling framework offers a principled approach to addressing several of these limitations, particularly the integration of preoperative and intraoperative predictors, interpretability, and calibration-aware evaluation. By organizing perioperative predictors into clinically coherent domains, such as patient vulnerability, procedural characteristics, and intraoperative physiology, domain-specific models can capture distinct risk mechanisms while preserving interpretability. Integrating domain-level predictions through ensemble learning enables complementary risk signals to be combined into a unified prediction, while allowing transparent attribution of risk across domains. When paired with rigorous calibration assessment and decision-analytic evaluation, such a framework supports translation from prediction to action~\cite{ref14,ref19}. Challenges such as generalizability across institutions and prospective validation remain and are addressed in our evaluation design and discussion.

In this work, we present a domain-structured ensemble framework for the development of perioperative prediction models using routinely collected electronic health record data from a statewide health information exchange. As a demonstration of this approach, we apply the framework to postoperative delirium (POD), organizing predictors into patient-related, surgery-related, and anesthetics-related domains, and integrating their predictions through a stacked meta-learner. Model performance is evaluated using patient-level grouped cross-validation, domain ablation analysis, and temporal validation, with explicit assessment of discrimination, calibration, and clinical utility, including adjustment for case--control sampling. Although illustrated in a single clinical application, this framework is designed to be extensible to diverse perioperative outcomes and to support interpretable, clinically actionable risk stratification across the perioperative continuum.

\section{Methods}

\subsection{Study Design and Data Sources}
We conducted a retrospective case--control study using electronic health record (EHR) data from the Indiana Network for Patient Care (INPC), a statewide health information exchange integrating patient-level clinical data from more than 100 healthcare systems across Indiana. The INPC captures inpatient and outpatient encounters, diagnoses, vital signs, laboratory results, medication administrations, and perioperative documentation, enabling longitudinal linkage across institutions.

The domain-structured framework is designed to be outcome-agnostic; while we demonstrate its application to postoperative delirium (POD), the architecture accommodates substitution of outcome definitions for other perioperative complications without modification to the modeling pipeline. Table~\ref{tab:domains} summarizes the mapping of predictor domains to representative clinical outcomes and provides the rationale for the three-domain organization.

\begin{table}[t]
\caption{Mapping of Predictor Domains to Clinical Outcomes}
\label{tab:domains}
\centering
\setlength{\tabcolsep}{2.5pt}
\renewcommand{\arraystretch}{1.15}
\scriptsize
\begin{tabular}{|p{0.16\columnwidth}|p{0.26\columnwidth}|p{0.26\columnwidth}|p{0.22\columnwidth}|}
\hline
\textbf{Domain} & \textbf{Representative Outcomes} & \textbf{Key Feature Types} & \textbf{Rationale} \\
\hline
Patient-related (baseline vulnerability) & POD; AKI; respiratory failure; prolonged LOS; mortality & Age; ASA status; CCI; chronic organ dysfunction; baseline medications & Captures physiologic reserve and susceptibility to stressors \\
\hline
Surgery-related (procedural stress) & POD; SSI; hemorrhage; ICU admission; reoperation & Procedure duration; emergency status; surgical specialty; complexity & Reflects magnitude and urgency of surgical insult \\
\hline
Anesthetics-related (physiologic perturbation) & POD; hypotension-related injury; myocardial injury; respiratory complications & MAP trends; vasopressor use; MAC-hours; neuromuscular blockade & Represents modifiable intraoperative exposures \\
\hline
\multicolumn{4}{|p{0.95\columnwidth}|}{\scriptsize POD, postoperative delirium; AKI, acute kidney injury; LOS, length of stay; SSI, surgical site infection; ICU, intensive care unit; ASA, American Society of Anesthesiologists; CCI, Charlson Comorbidity Index; MAP, mean arterial pressure; MAC, minimum alveolar concentration.} \\
\hline
\end{tabular}
\end{table}

\subsection{Study Population}
Adults aged $\geq$18 years who underwent non-cardiac, non-obstetric inpatient surgery between 2010 and 2021 were eligible. Patients required perioperative EHR data and at least one Confusion Assessment Method (CAM) evaluation within seven postoperative days. Patients with preexisting dementia (ICD-9: 290.x, 294.1x, 331.0--331.2; ICD-10: F01--F03, G30, G31.0--G31.2) were excluded to isolate incident postoperative delirium from chronic cognitive impairment, consistent with standard POD prediction methodology~\cite{ref19}.

For analytic efficiency, we employed 1:1 matching of outcome-positive to outcome-negative encounters on age ($\pm$5 years), sex, race, and year of surgery. The implications of case--control sampling for calibration and clinical utility are addressed in subsequent sections.

\subsection{Outcome Definition}
Postoperative delirium was identified using dual confirmation: (1) delirium-related ICD codes and (2) positive CAM assessment, both within seven postoperative days. The ICD code set encompassed $>$80 codes spanning delirium presentations including physiological delirium (F05, 293.0--293.1), encephalopathy (G93.40, R41.82), drug-induced delirium (F11--F19.x21), alcohol withdrawal delirium (F10.121, 291.0), and altered mental status (R41.0, 780.97). The CAM has reported sensitivity of 94--100\% and specificity of 90--95\%~\cite{ref20}.

Controls had at least one documented CAM assessment with negative result and no delirium-related ICD codes during the seven-day window. Table~\ref{tab:outcome} summarizes the outcome definition framework.

\begin{table}[t]
\caption{Outcome Definition Framework}
\label{tab:outcome}
\centering
\setlength{\tabcolsep}{4pt}
\renewcommand{\arraystretch}{1.2}
\footnotesize
\begin{tabular}{|p{0.18\columnwidth}|p{0.74\columnwidth}|}
\hline
\textbf{Element} & \textbf{Specification} \\
\hline
\multicolumn{2}{|l|}{\textit{Exemplar Application: Postoperative Delirium}} \\
\hline
Case definition & Dual confirmation: delirium-related ICD code(s) \textbf{AND} positive CAM, both within 7 postoperative days \\
\hline
Control definition & Documented CAM with negative result \textbf{AND} no delirium ICD codes during 7-day window \\
\hline
Exclusion & Preexisting dementia diagnosis \\
\hline
\multicolumn{2}{|l|}{\textit{Framework Extensibility}} \\
\hline
Alternative outcomes & AKI, respiratory failure, SSI, ICU admission, 30-day mortality \\
\hline
Adaptation & Substitute ICD codes and validation criteria; no pipeline modification required \\
\hline
\end{tabular}
\end{table}

\subsection{Predictor Variables}
Candidate predictors were extracted from structured EHR data and organized into three domains (Table~\ref{tab:domains}). An initial 104 variables were reduced to 74 after filtering pairs with PhiK correlation $>$0.8~\cite{ref21}. The ASA classification was collapsed from 12 dummies to two features: numeric class (1--6) and emergency flag.

The final predictor set comprised: \textbf{patient-related} ($n=31$): demographics, ASA class, Charlson Comorbidity Index, 22 comorbidity indicators, and 4 medication classes; \textbf{surgery-related} ($n=12$): duration, emergency status, and 10 surgical service categories; and \textbf{anesthetics-related} ($n=31$): MAC-hours, state entropy, 13 anesthetic agents, 10 vasoactive medications, and 6 MAP-derived hemodynamic measures.

Baseline characteristics stratified by outcome status are presented in Table~\ref{tab:baseline}.

\begin{table}[t]
\caption{Baseline Characteristics of the Study Cohort}
\label{tab:baseline}
\centering
\setlength{\tabcolsep}{3pt}
\renewcommand{\arraystretch}{1.1}
\footnotesize
\begin{tabular}{|p{0.45\columnwidth}|c|c|}
\hline
\textbf{Characteristic} & \textbf{\makecell{Outcome-\\Negative}} & \textbf{\makecell{Outcome-\\Positive}} \\
 & \textbf{(n=2,693)} & \textbf{(n=2,693)} \\
\hline
\multicolumn{3}{|l|}{\textit{Demographics}\textsuperscript{a}} \\
\hline
Age, years & 65 (52--75) & 65 (52--75) \\
Female sex & 1,303 (48.4) & 1,303 (48.4) \\
White race & 2,390 (88.7) & 2,390 (88.7) \\
\hline
\multicolumn{3}{|l|}{\textit{Clinical Characteristics}} \\
\hline
ASA physical status & 3 (2--3) & 3 (3--4) \\
Emergency surgery & 157 (5.8) & 373 (13.9) \\
Charlson Comorbidity Index & 1 (0--2) & 1 (0--3) \\
\hline
\multicolumn{3}{|l|}{\textit{Surgical Characteristics}} \\
\hline
Surgical duration, min & 43 (20--86) & 76 (37--146) \\
Neurosurgery & 171 (6.3) & 461 (17.1) \\
Ophthalmologic surgery & 266 (9.9) & 11 (0.4) \\
\hline
\multicolumn{3}{|l|}{\textit{Intraoperative Hemodynamics}} \\
\hline
Minimum MAP, mmHg & 62.3 (54.7--72.3) & 56.3 (49.0--64.0) \\
Time MAP $<$65 mmHg, min & 3 (0--18) & 12 (2--36) \\
\hline
\multicolumn{3}{|l|}{\textit{Anesthetic Exposures}} \\
\hline
Total MAC-hours & 0.67 (0--1.60) & 1.11 (0.36--2.21) \\
Phenylephrine & 573 (21.3) & 1,160 (43.1) \\
\hline
\multicolumn{3}{|l|}{\textit{Selected Comorbidities}} \\
\hline
Psychiatric disorders & 895 (33.2) & 1,227 (45.6) \\
Respiratory failure & 288 (10.7) & 714 (26.5) \\
Cardiac arrhythmia & 843 (31.3) & 1,078 (40.0) \\
\hline
\multicolumn{3}{|p{0.95\columnwidth}|}{\scriptsize\textsuperscript{a}Matched 1:1 by age, sex, race, and year. Continuous: median (IQR); categorical: n (\%). Data shown for POD exemplar.} \\
\hline
\end{tabular}
\end{table}

\subsection{Data Preprocessing}
Missingness was minimal (ethnicity 0.5\%, ASA $<$0.1\%). Continuous variables were median-imputed and robust-scaled; categorical variables were mode-imputed and one-hot encoded; binary variables were mode-imputed only. All preprocessing was implemented in scikit-learn pipelines maintained in an unfitted state, fit exclusively on training folds to prevent information leakage.

\subsection{Sample Size}
Sample size was estimated using the Riley--van Smeden formula~\cite{ref18}:
\begin{equation}
n = \exp\!\left(
\frac{-0.508 + 0.259 \ln(\phi) + 0.504 \ln(p) - \ln(m)}{0.544}
\right)
\end{equation}
where $\phi$ is outcome proportion, $p$ is number of predictors, and $m$ is target mean absolute prediction error. With $\phi=0.5$, $p=74$, and $m=0.05$, minimum required sample was 3,756 encounters. The final cohort of 5,386 encounters (2,693 per group) exceeded this by 43\%.

\subsection{Model Development}
Model development proceeded in three stages (Table~\ref{tab:models}). All models used stratified 5-fold cross-validation with patient-level grouping.

\textbf{Baseline models}: Four classifiers were trained on all 74 predictors: $\ell_2$ logistic regression, elastic net logistic regression, random forest (500 trees), and histogram gradient boosting.

\textbf{Tuned gradient boosting}: XGBoost was tuned via nested cross-validation (5-fold outer, 4-fold inner) with light (40 iterations) and expanded (60 iterations with isotonic calibration) protocols.

\textbf{Domain-structured ensemble}: Three LightGBM classifiers were trained independently on domain-specific predictor subsets (patient: 31, surgery: 12, anesthetics: 31 features) with shared hyperparameters (600 trees, learning rate 0.03, 31 leaves). Out-of-fold probabilities from domain models were combined via a logistic regression meta-learner ($\ell_2$, C=1.0), whose coefficients quantify relative domain contributions.

\begin{table}[t]
\caption{Model Specifications}
\label{tab:models}
\centering
\setlength{\tabcolsep}{2.5pt}
\renewcommand{\arraystretch}{1.1}
\footnotesize
\begin{tabular}{|p{0.28\columnwidth}|p{0.65\columnwidth}|}
\hline
\textbf{Model} & \textbf{Key Parameters} \\
\hline
\multicolumn{2}{|l|}{\textit{Baseline Models}} \\
\hline
Logistic regression ($\ell_2$) & C=1.0; solver=liblinear; max\_iter=5,000 \\
Elastic net & $\ell_1$ ratio=0.2; C=1.0; solver=saga \\
Random forest & 500 trees; min\_samples\_split=4; min\_samples\_leaf=2 \\
Histogram gradient boosting & Learning rate=0.08; max\_bins=255; early stopping \\
\hline
\multicolumn{2}{|l|}{\textit{Tuned XGBoost}} \\
\hline
Light / Expanded & 800--1,200 trees; 40--60 iterations; depth 3--7 \\
\hline
\multicolumn{2}{|l|}{\textit{Domain-Specific LightGBM}} \\
\hline
All domains & 600 trees; LR=0.03; 31 leaves; min\_samples=40 \\
\hline
\multicolumn{2}{|l|}{\textit{Meta-Learner}} \\
\hline
Logistic regression & $\ell_2$; C=1.0; 3 input features (domain probabilities) \\
\hline
\multicolumn{2}{|p{0.93\columnwidth}|}{\scriptsize 5-fold patient-grouped CV. XGBoost: nested CV (4-fold inner). Meta-learner trained on OOF domain probabilities.} \\
\hline
\end{tabular}
\end{table}

\subsection{Domain Ablation and Sensitivity Analyses}
To assess whether the three-domain architecture provides meaningful benefit beyond a surgery-only model, we conducted a systematic domain ablation study. Seven configurations were evaluated: the full three-domain ensemble, three two-domain combinations (dropping one domain each), and three single-domain models. All configurations used identical cross-validation and meta-learner procedures.

Sensitivity analyses assessed robustness to: (1) meta-learner architecture (logistic regression, unpenalized logistic regression, random forest, gradient boosting); (2) case--control ratio (1:1, 1:2, 1:3 via case downsampling); and (3) cross-validation scheme (5-fold vs.\ 10-fold grouped CV). Bootstrap confidence intervals (1,000 resamples) were computed for primary metrics.

\subsection{Temporal Validation}
To evaluate generalizability across time periods, we performed temporal validation by training the full domain ensemble on encounters from 2010--2017 ($n=3{,}237$) and testing on 2018--2021 ($n=2{,}149$). Within the training set, out-of-fold domain predictions were generated via 5-fold GroupKFold to train the meta-learner, preventing calibration leakage. Domain models were then refit on the full training set for test-set inference. Temporal cutoff sensitivity was assessed at 2016, 2017, 2018, and 2019 boundaries.

\subsection{Model Interpretability}
Feature contributions were quantified using SHAP~\cite{ref22} for each domain-specific model separately, preserving clinically meaningful attribution within coherent predictor groups. Domain models were refit on the full dataset; TreeExplainer provided exact Shapley values. Global importance was summarized by mean absolute SHAP value; beeswarm plots visualized feature effect distributions.

\subsection{Model Calibration and Clinical Utility}
Calibration was assessed using aggregated out-of-fold predictions. Calibration curves were constructed by decile; intercept (ideal=0) and slope (ideal=1) were estimated with 95\% bootstrap CIs (1,000 resamples).

Because case--control sampling yields artificial 50\% prevalence, predicted probabilities were adjusted for target populations using:
\begin{equation}
\text{logit}(p_{\text{adj}}) = \text{logit}(p) + \text{logit}(\pi_{\text{target}}) - \text{logit}(\pi_{\text{source}})
\end{equation}
Calibration was recomputed at prevalences of 10\%, 20\%, and 40\%.

Clinical utility was evaluated via decision curve analysis (DCA)~\cite{ref14} with inverse-probability weighting to account for case--control sampling~\cite{ref23}. Net benefit was calculated across threshold probabilities 0.05--0.50 for each target prevalence.

Performance was summarized using AUROC, PR-AUC, and Brier score from aggregated OOF predictions, with per-fold SDs reported.

\subsection{Ethics and Software}
This study was deemed exempt by the Indiana University IRB (Protocol \#13577). Analyses used Python 3.9 with scikit-learn (v1.3), XGBoost (v1.7), LightGBM (v4.0), SHAP (v0.42), and statsmodels (v0.14). Code is available at \\ {\small\url{https://github.com/Amorfati123/periop-prediction-framework}}.

\section{Results}

\subsection{Baseline and Tuned Model Performance}
Performance of baseline models and tuned XGBoost classifiers is summarized in Table~\ref{tab:model_performance}. Among baseline approaches trained on all 74 predictors, histogram-based gradient boosting achieved the highest discrimination (AUROC 0.849, PR-AUC 0.832) and lowest Brier score (0.158). Logistic regression and random forest models yielded moderate discrimination (AUROC 0.809--0.821). Performance was stable across folds (SD $<$ 0.01 for AUROC).

Hyperparameter tuning of XGBoost via nested cross-validation did not improve upon baseline performance. Light tuning achieved AUROC 0.827; expanded tuning with isotonic calibration reached AUROC 0.839, both below the histogram gradient boosting baseline. These findings motivated exploration of domain-structured architectures.

\begin{table}[t]
\caption{Model Performance Comparison}
\label{tab:model_performance}
\centering
\setlength{\tabcolsep}{4pt}
\renewcommand{\arraystretch}{1.15}
\footnotesize
\begin{tabular}{|p{0.38\columnwidth}|c|c|c|}
\hline
\textbf{Model} & \textbf{AUROC} & \textbf{PR-AUC} & \textbf{Brier} \\
\hline
\multicolumn{4}{|l|}{\textit{Baseline Models (74 predictors)}} \\
\hline
Logistic regression ($\ell_2$) & 0.813 & 0.799 & 0.177 \\
Elastic net & 0.809 & 0.795 & 0.179 \\
Random forest & 0.821 & 0.810 & 0.175 \\
Histogram gradient boosting & 0.849 & 0.832 & 0.158 \\
\hline
\multicolumn{4}{|l|}{\textit{Tuned XGBoost}} \\
\hline
Light tuning & 0.827 & 0.816 & 0.169 \\
Expanded + isotonic & 0.839 & 0.829 & 0.163 \\
\hline
\multicolumn{4}{|l|}{\textit{Domain-Specific Models}} \\
\hline
Patient-related & 0.735 $\pm$ 0.017 & 0.729 $\pm$ 0.022 & 0.212 \\
Surgery-related & 0.879 $\pm$ 0.010 & 0.859 $\pm$ 0.017 & 0.140 \\
Anesthetics-related & 0.720 $\pm$ 0.009 & 0.720 $\pm$ 0.015 & 0.219 \\
\hline
\multicolumn{4}{|l|}{\textit{Stacked Meta-Learner}} \\
\hline
Combined (3 domains) & \textbf{0.899} & \textbf{0.881} & \textbf{0.126} \\
\hline
\multicolumn{4}{|p{0.93\columnwidth}|}{\scriptsize All metrics from aggregated OOF predictions (5-fold patient-grouped CV). Domain models: mean $\pm$ SD across folds. Meta-learner 95\% bootstrap CIs: AUROC (0.891, 0.906); PR-AUC (0.869, 0.892); Brier (0.121, 0.132).} \\
\hline
\end{tabular}
\end{table}

\subsection{Domain-Structured Ensemble Performance}
Among individual domain models, the surgery-related model demonstrated strongest standalone discrimination (AUROC 0.879), followed by patient-related (0.735) and anesthetics-related (0.720) domains (Table~\ref{tab:model_performance}). The stacked meta-learner achieved AUROC of 0.899 (95\% CI: 0.891--0.906), PR-AUC of 0.881, and Brier score of 0.126, exceeding the best baseline model by 0.050 in AUROC.

Meta-learner coefficients indicated that surgery-related predictions contributed 96.7\% of signal to the final risk estimate, followed by patient-related (2.7\%) and anesthetics-related (0.6\%) domains. These weightings are specific to the POD exemplar and reflect the dominant role of procedural stress in delirium pathophysiology; alternative outcomes would yield different domain contributions.

\subsection{Domain Ablation Study}
Table~\ref{tab:ablation} presents the systematic ablation results. The full three-domain ensemble (AUROC 0.899, Brier 0.126) outperformed the surgery-only model (AUROC 0.879, Brier 0.140), yielding a 0.020 AUROC improvement and a 10\% reduction in Brier score. Dropping the patient-related domain reduced AUROC to 0.883 and increased Brier to 0.136, indicating that patient-related features contribute meaningfully to calibration despite their modest meta-learner coefficient. Dropping the anesthetics-related domain had minimal impact (AUROC 0.898), while dropping the surgery-related domain caused severe degradation (AUROC 0.779). Notably, the surgery-related domain contains only 12 features compared with 31 each for patient-related and anesthetics-related domains, confirming that its dominance reflects genuine clinical signal rather than feature space size imbalance.

\begin{table}[t]
\caption{Domain Ablation Study Results}
\label{tab:ablation}
\centering
\setlength{\tabcolsep}{2.5pt}
\renewcommand{\arraystretch}{1.15}
\footnotesize
\begin{tabular}{|p{0.38\columnwidth}|c|c|c|c|}
\hline
\textbf{Configuration} & \textbf{$k$} & \textbf{AUROC} & \textbf{PR-AUC} & \textbf{Brier} \\
\hline
All 3 domains & 3 & \textbf{0.899} & \textbf{0.880} & \textbf{0.126} \\
Drop anesthetics & 2 & 0.898 & 0.881 & 0.128 \\
Drop patient & 2 & 0.883 & 0.859 & 0.136 \\
Surgery only & 1 & 0.879 & 0.857 & 0.140 \\
Drop surgery & 2 & 0.779 & 0.767 & 0.192 \\
Patient only & 1 & 0.734 & 0.725 & 0.209 \\
Anesthetics only & 1 & 0.720 & 0.718 & 0.214 \\
\hline
\multicolumn{5}{|p{0.93\columnwidth}|}{\scriptsize $k$, number of active domains. All metrics from aggregated OOF predictions (5-fold patient-grouped CV).} \\
\hline
\end{tabular}
\end{table}

\subsection{Temporal Validation}
In temporal validation (train $<$2018, test $\geq$2018), the meta-learner achieved AUROC of 0.915, PR-AUC of 0.902, and Brier score of 0.118 on the held-out test set ($n=2{,}149$). Domain-level performance on the test set was consistent with cross-validated estimates: surgery-related AUROC 0.896, patient-related 0.748, anesthetics-related 0.738.

Calibration on the temporal test set showed intercept 0.139 (95\% CI: 0.014, 0.264) and slope 1.330 (95\% CI: 1.224, 1.437). The slope exceeding 1.0 indicates mild underconfidence in predicted probabilities, a pattern commonly observed in temporal validation and addressable through recalibration at deployment. Temporal cutoff sensitivity analysis showed stable discrimination across boundaries: AUROC 0.858 (2016 cutoff), 0.894 (2017), 0.915 (2018), and 0.913 (2019), with performance improving monotonically with training set size.

\subsection{Sensitivity Analyses}
Meta-learner architecture had minimal impact on performance: logistic regression (AUROC 0.899), unpenalized logistic regression (0.899), and gradient boosting (0.899) all performed equivalently; random forest meta-learner was slightly lower (0.887). Simulated case--control ratios of 1:2 and 1:3 yielded AUROC of 0.887 and 0.867, respectively, demonstrating graceful degradation rather than dependence on balanced sampling. Cross-validation scheme (5-fold vs.\ 10-fold) produced equivalent estimates (AUROC 0.899 vs.\ 0.898). Subgroup analysis showed consistent performance across sex (AUC gap 0.005), age groups (gap 0.048), and race (gap 0.073).

\subsection{Feature-Level Interpretability}
\begin{figure}[t]
\centering
\includegraphics[width=\columnwidth]{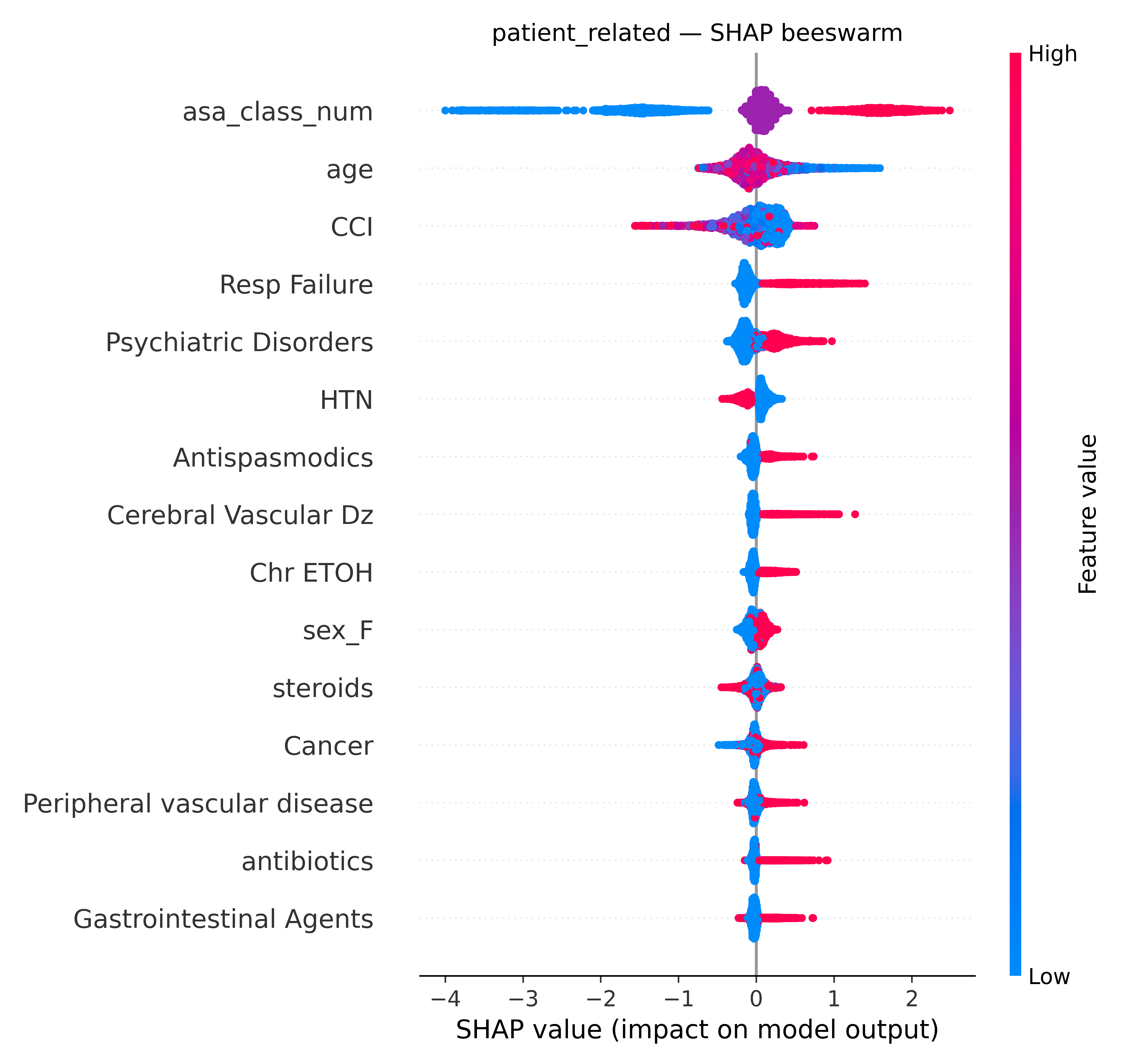}
\caption{SHAP beeswarm plot for patient-related domain. Color indicates feature value (red = high, blue = low); horizontal position indicates contribution to predicted risk. Variable abbreviations: asa\_class\_num, ASA physical status (numeric); CCI, Charlson Comorbidity Index; Resp Failure, respiratory failure history; HTN, hypertension; Cerebral Vascular Dz, cerebrovascular disease; Chr ETOH, chronic alcohol use; sex\_F, female sex.}
\label{fig:shap_patient}
\end{figure}

\begin{figure}[t]
\centering
\includegraphics[width=\columnwidth]{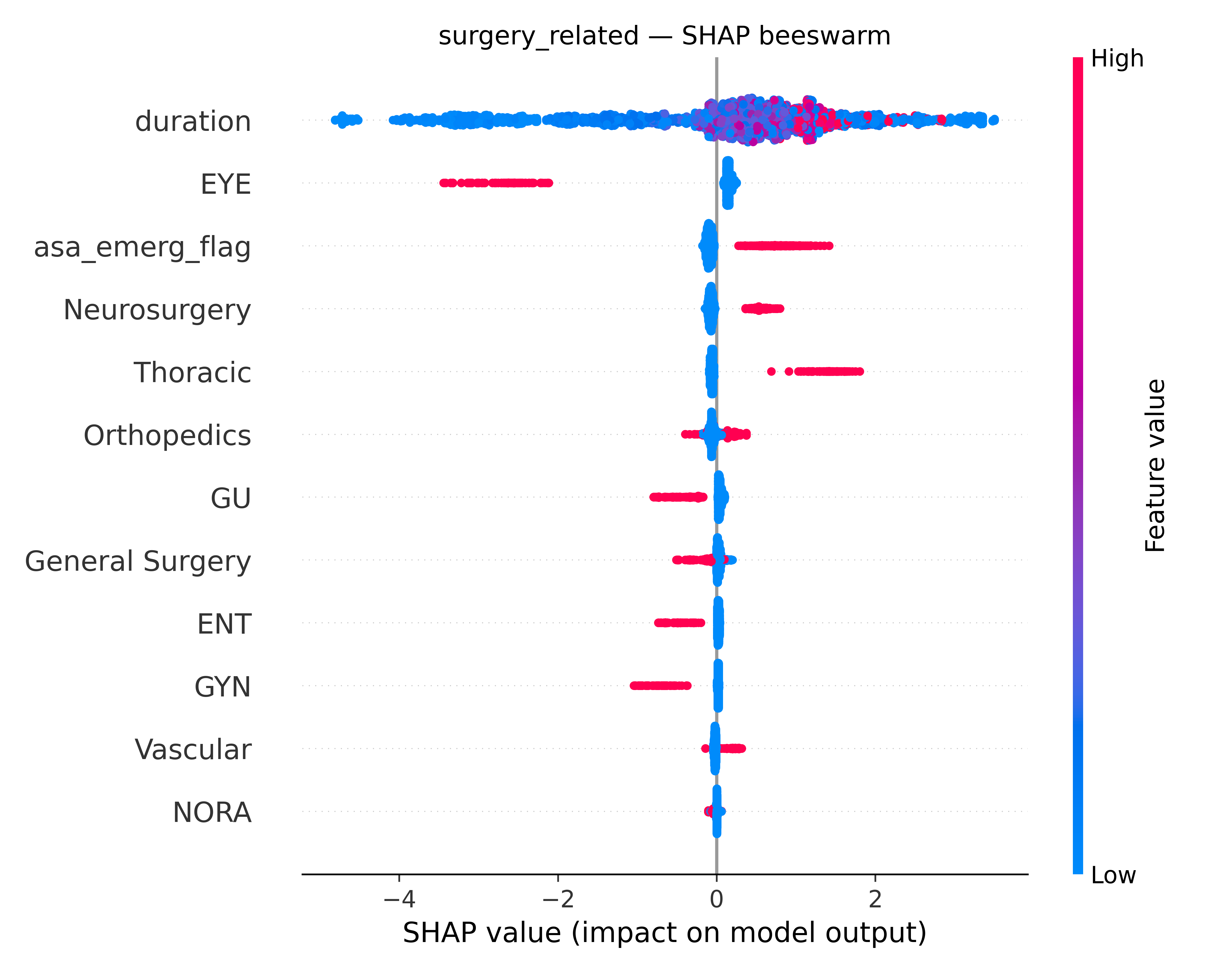}
\caption{SHAP beeswarm plot for surgery-related domain. Operative duration is dominant; ophthalmologic surgery (EYE) shows strong protective effect. Variable abbreviations: asa\_emerg\_flag, emergency surgical status; GU, genitourinary surgery; ENT, ear-nose-throat surgery; GYN, gynecologic surgery; NORA, non-operating room anesthesia.}
\label{fig:shap_surgery}
\end{figure}

\begin{figure}[t]
\centering
\includegraphics[width=\columnwidth]{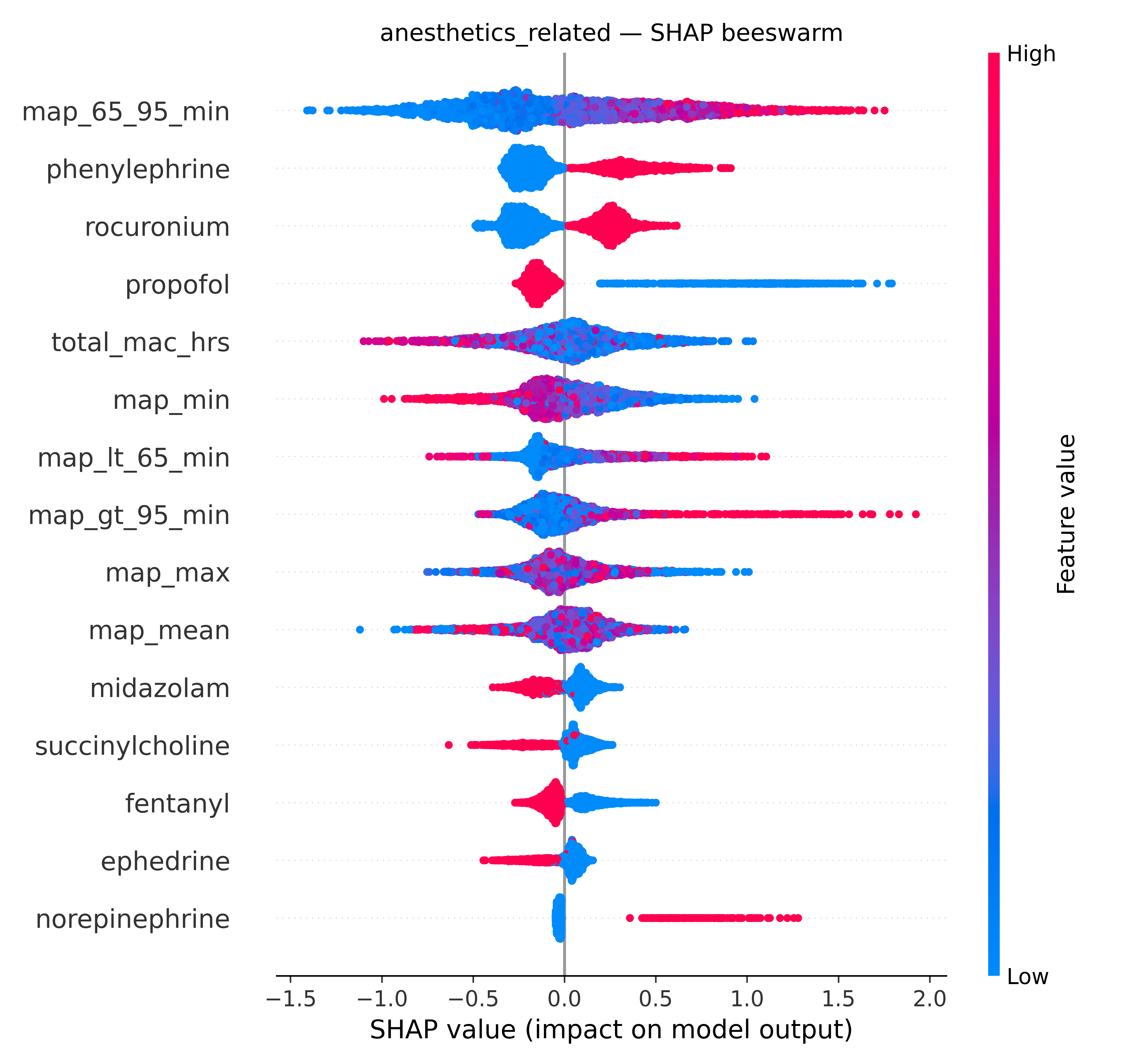}
\caption{SHAP beeswarm plot for anesthetics-related domain. Normotensive duration is the leading contributor. Variable abbreviations: map\_65\_95\_min, time with MAP 65--95 mmHg (min); total\_mac\_hrs, total minimum alveolar concentration hours; map\_min, minimum MAP; map\_lt\_65\_min, time with MAP $<$65 mmHg (min); map\_gt\_95\_min, time with MAP $>$95 mmHg (min); map\_max, maximum MAP; map\_mean, mean MAP. All MAP values in mmHg.}
\label{fig:shap_anesthetics}
\end{figure}

Feature contributions were assessed using SHAP values for each domain model (Table~\ref{tab:shap}; Figures~\ref{fig:shap_patient}--\ref{fig:shap_anesthetics}).

In the surgery-related domain, operative duration was dominant (mean $|$SHAP$|$ = 1.31), with longer procedures increasing risk. Ophthalmologic surgery showed protective effects (0.28); emergency status (0.16) and neurosurgery (0.13) increased risk.

For patient-related features, ASA physical status led (0.79), followed by age (0.22), Charlson Comorbidity Index (0.22), respiratory failure (0.21), and psychiatric disorders (0.18), consistent with established POD risk factors.

In the anesthetics-related domain, time with MAP 65--95 mmHg was most important (0.43), with longer normotensive duration reducing risk. Phenylephrine (0.25), rocuronium (0.24), propofol (0.24), and MAC-hours (0.22) also contributed substantially.

\begin{table}[t]
\caption{Top Predictors by Domain (SHAP Importance)}
\label{tab:shap}
\centering
\setlength{\tabcolsep}{4pt}
\renewcommand{\arraystretch}{1.15}
\footnotesize
\begin{tabular}{|p{0.28\columnwidth}|p{0.45\columnwidth}|c|}
\hline
\textbf{Domain} & \textbf{Feature} & \textbf{$|$SHAP$|$} \\
\hline
\multicolumn{3}{|l|}{\textit{Surgery-related}} \\
\hline
 & Operative duration & 1.309 \\
 & Ophthalmologic surgery & 0.285 \\
 & Emergency status & 0.161 \\
\hline
\multicolumn{3}{|l|}{\textit{Patient-related}} \\
\hline
 & ASA physical status & 0.788 \\
 & Age & 0.220 \\
 & Charlson Comorbidity Index & 0.219 \\
 & Respiratory failure & 0.211 \\
\hline
\multicolumn{3}{|l|}{\textit{Anesthetics-related}} \\
\hline
 & Time MAP 65--95 mmHg & 0.430 \\
 & Phenylephrine & 0.252 \\
 & Rocuronium & 0.243 \\
\hline
\multicolumn{3}{|p{0.93\columnwidth}|}{\scriptsize Top 3--4 features per domain shown. Full rankings in beeswarm plots (Figures~\ref{fig:shap_patient}--\ref{fig:shap_anesthetics}).} \\
\hline
\end{tabular}
\end{table}

\subsection{Model Calibration and Clinical Utility}
\begin{figure}[t]
\centering
\includegraphics[width=0.85\columnwidth]{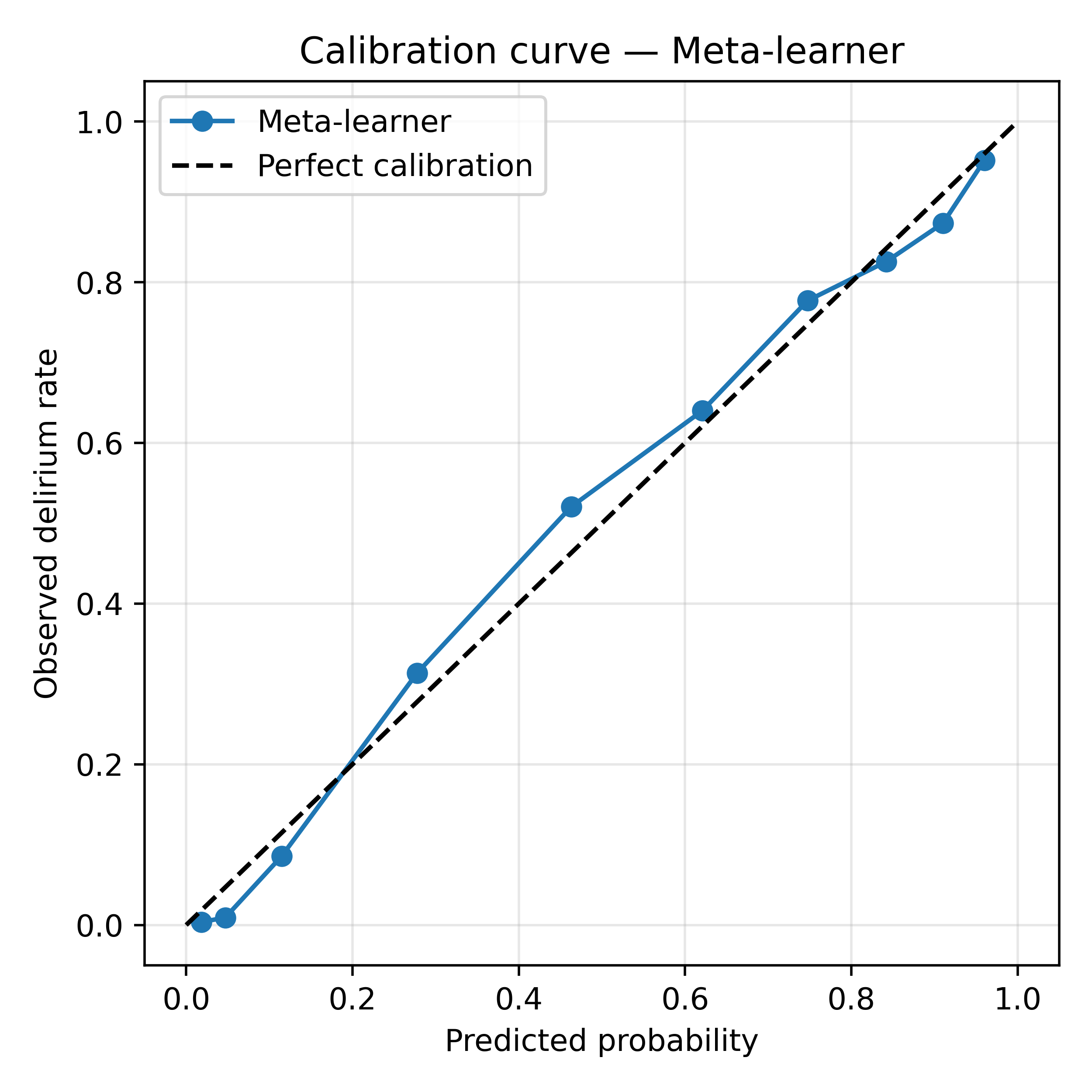}
\caption{Calibration curve for the stacked meta-learner showing observed vs.\ predicted outcome rates by decile. Dashed line indicates perfect calibration.}
\label{fig:calibration}
\end{figure}

\begin{figure}[t]
\centering
\includegraphics[width=0.85\columnwidth]{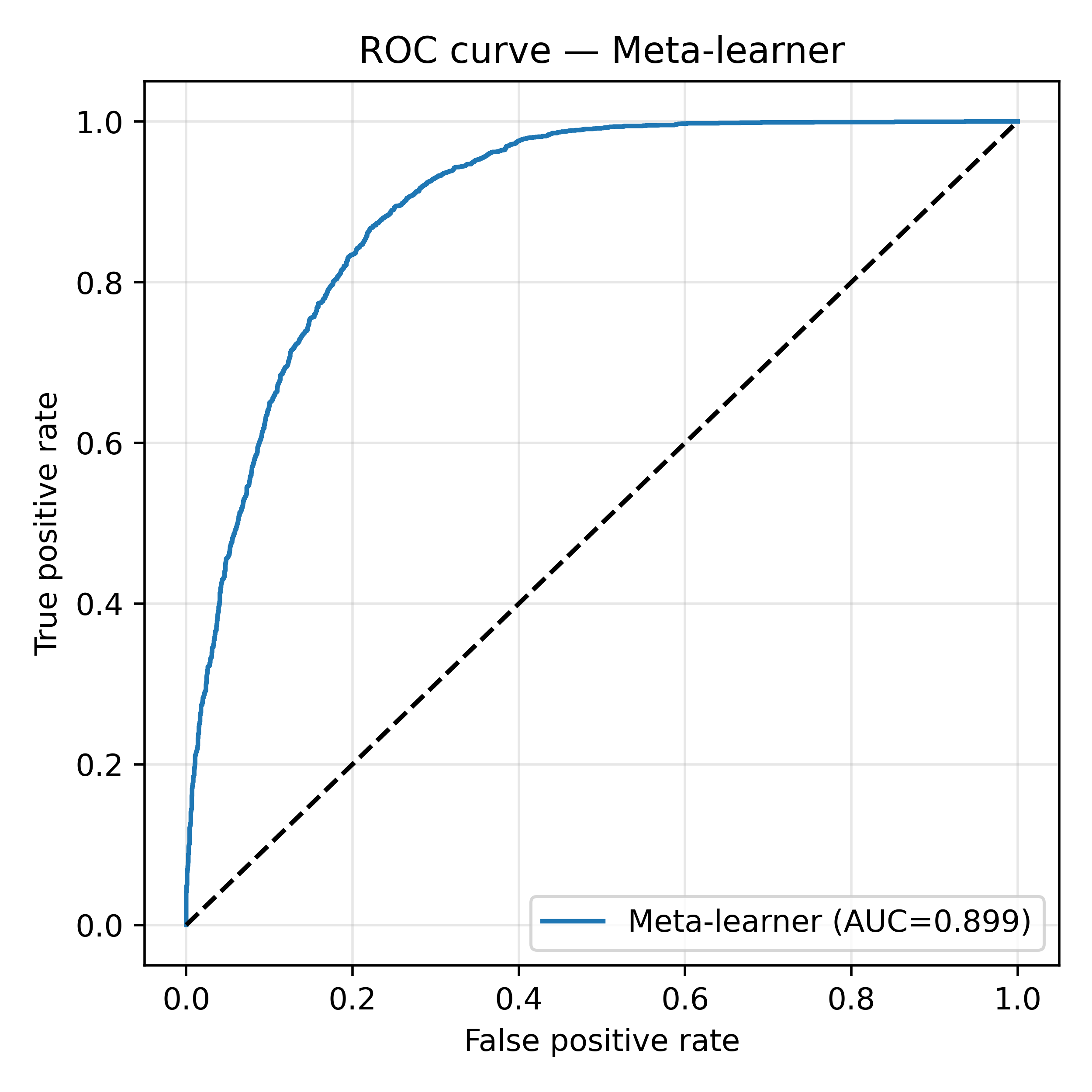}
\caption{ROC curve for the stacked meta-learner (AUROC = 0.899).}
\label{fig:roc}
\end{figure}

The meta-learner demonstrated excellent calibration (Figure~\ref{fig:calibration}): intercept $-0.006$ (95\% CI: $-0.083$ to 0.070) and slope 1.035 (95\% CI: 0.982 to 1.088), with confidence intervals encompassing ideal values (Table~\ref{tab:calibration}). After prevalence adjustment to target rates of 10\%, 20\%, and 40\%, slope remained unchanged while intercept shifted appropriately. The ROC curve (Figure~\ref{fig:roc}) confirmed strong discrimination (AUROC 0.899).

\begin{figure}[t]
\centering
\includegraphics[width=0.85\columnwidth]{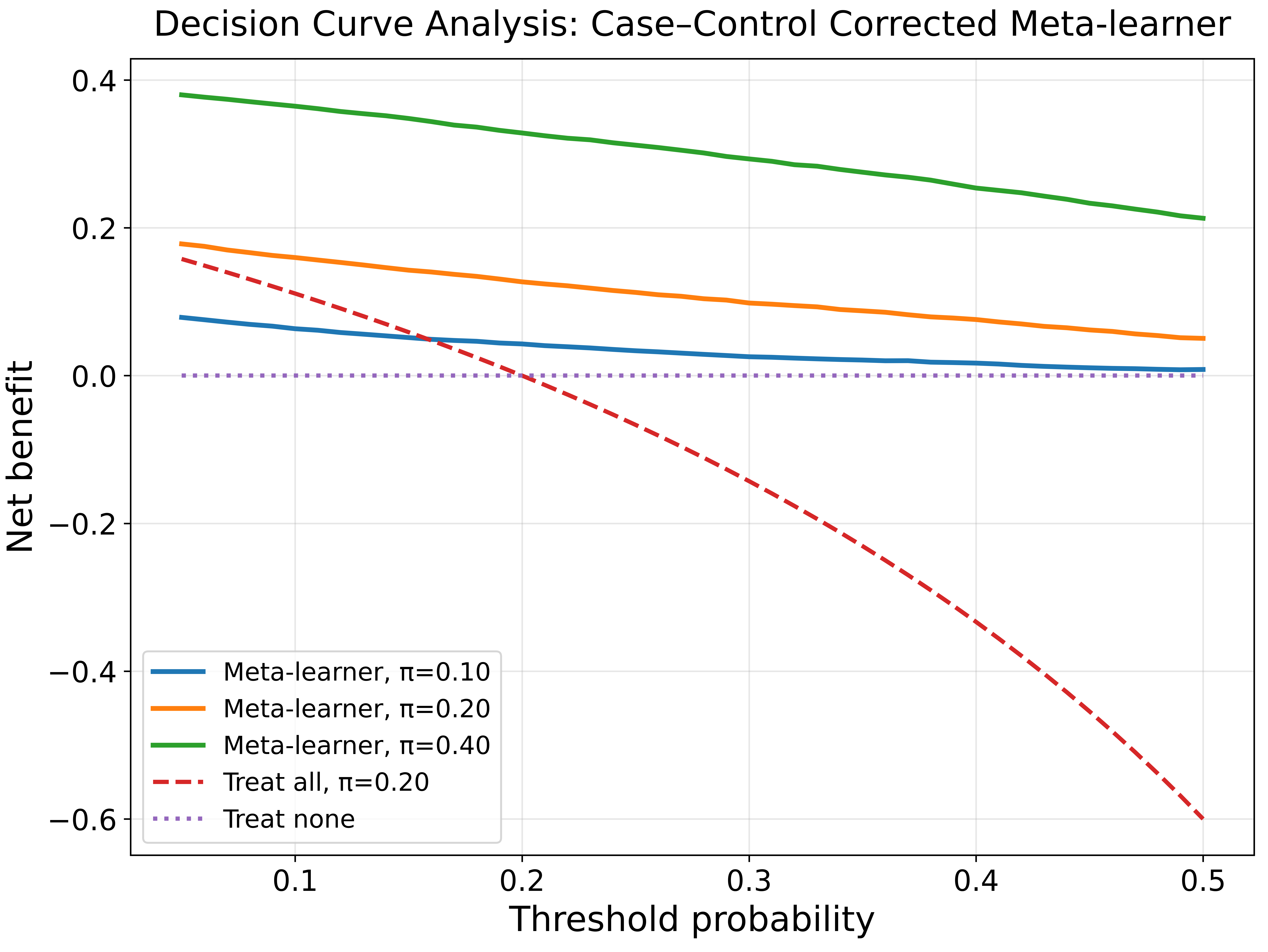}
\caption{Decision curve analysis showing net benefit across threshold probabilities for target prevalences of 10\%, 20\%, and 40\%.}
\label{fig:dca}
\end{figure}

Decision curve analysis with inverse-probability weighting (Figure~\ref{fig:dca}) demonstrated positive net benefit compared with default strategies across threshold probabilities 0.05--0.50 for all target prevalences. Net benefit was greatest at lower thresholds and higher prevalences, supporting the utility of framework-guided risk stratification for targeted intervention.

\begin{table}[t]
\caption{Calibration Metrics}
\label{tab:calibration}
\centering
\setlength{\tabcolsep}{3pt}
\renewcommand{\arraystretch}{1.15}
\footnotesize
\begin{tabular}{|p{0.28\columnwidth}|c|c|c|}
\hline
\textbf{Scenario} & \textbf{$\pi$} & \textbf{Intercept (95\% CI)} & \textbf{Slope (95\% CI)} \\
\hline
Case--control & 0.50 & $-0.006$ ($-0.083$, 0.070) & 1.035 (0.982, 1.088) \\
Adjusted & 0.10 & 2.268 (2.137, 2.399) & 1.035 (0.982, 1.088) \\
Adjusted & 0.20 & 1.428 (1.329, 1.528) & 1.035 (0.982, 1.088) \\
Adjusted & 0.40 & 0.413 (0.337, 0.490) & 1.035 (0.982, 1.088) \\
\hline
\multicolumn{4}{|p{0.93\columnwidth}|}{\scriptsize $\pi$, target prevalence. Ideal: intercept = 0, slope = 1. Slope invariant to prevalence adjustment.} \\
\hline
\end{tabular}
\end{table}

\section{Discussion}

This work presents a domain-structured ensemble framework for perioperative risk prediction using routinely collected electronic health record data. By organizing predictors into three clinically motivated domains, namely patient-related baseline vulnerability, surgery-related procedural characteristics, and anesthetics-related intraoperative exposures, the framework reflects the multifactorial pathophysiology underlying perioperative complications. Applied to postoperative delirium as an exemplar, the stacked meta-learner achieved AUROC of 0.899 (95\% CI: 0.891--0.906), exceeding the best monolithic baseline by 0.050, with excellent calibration and positive net benefit across clinically relevant decision thresholds. Temporal validation on held-out post-2017 data demonstrated maintained discrimination (AUROC 0.915), and domain ablation analysis confirmed that the multi-domain architecture improves both discrimination and probability estimation relative to any single-domain model.

\subsection{Comparison With Existing Approaches}
The framework addresses key limitations of existing perioperative risk prediction tools. Traditional scores such as ASA classification and RCRI rely primarily on preoperative factors and offer limited discriminatory power~\cite{ref12,ref13}. Procedure-specific models improve discrimination but lack generalizability across surgical contexts. Contemporary machine learning approaches often incorporate intraoperative data but treat predictors as a flat feature space, limiting interpretability and modularity~\cite{ref9}. In contrast, the proposed framework decomposes prediction into domain-specific submodels trained independently with standardized pipelines, followed by a meta-learner that integrates domain-level risk estimates. This architecture enables modular development, supports substitution or extension of domain models, and preserves interpretability while improving predictive performance.

\subsection{Clinical and Methodological Implications}
The POD exemplar illustrates the framework's utility. Surgery-related features contributed the largest share of predictive signal (96.7\%), reflecting the dominant role of procedural complexity and duration in perioperative stress. The domain ablation study (Table~\ref{tab:ablation}) provides direct evidence that this dominance does not render the three-domain architecture unnecessary. The full ensemble improved Brier score from 0.140 (surgery-only) to 0.126 (three-domain), a 10\% reduction indicating meaningfully better probability estimation. Although the patient-related domain contributed only 2.7\% of meta-learner signal by coefficient magnitude, removing it degraded AUROC from 0.899 to 0.883 and Brier from 0.126 to 0.136. These improvements in calibration and probability accuracy are clinically important because decision support tools require well-calibrated risk estimates, not merely correct rankings~\cite{ref14,ref15}.

Notably, the surgery-related domain contains only 12 features compared with 31 each for the patient-related and anesthetics-related domains. Its predictive dominance therefore reflects genuine concentration of POD-relevant clinical signal in procedural variables, not an artifact of feature space size. Importantly, domain contributions are learned empirically through stacking rather than assumed a priori, a property critical for generalization, as different outcomes may be driven by different combinations of vulnerability, procedural stress, and intraoperative exposure.

Meta-learner choice had negligible impact on performance: logistic regression, unpenalized logistic regression, and gradient boosting all yielded equivalent AUROC (0.899), confirming that the stacking benefit derives from domain-level signal decomposition rather than meta-learner complexity.

Feature-level interpretability was preserved through SHAP analyses applied to domain-specific models rather than the ensemble, enabling clinically coherent risk attribution while avoiding the opacity of monolithic ensembles. Calibration assessment and decision curve analysis, implemented as framework-level evaluation components, ensure that predicted probabilities support clinical decision-making rather than ranking alone, addressing a common limitation of existing perioperative prediction systems~\cite{ref14,ref15}.

\subsection{Framework Extensibility}
A key advantage is extensibility across perioperative outcomes. As outlined in Table~\ref{tab:domains}, the three-domain structure maps naturally to diverse endpoints including acute kidney injury, respiratory complications, surgical site infection, ICU admission, and mortality. For each outcome, the framework requires modification only to case and control labeling functions; the modeling architecture, cross-validation strategy, stacking logic, and evaluation procedures remain unchanged. This separation of outcome definition from modeling infrastructure enables consistent development across multiple endpoints while minimizing engineering overhead.

The framework also supports dynamic risk updating. Because domain-level predictions are generated independently, risk estimates can be updated as additional perioperative data become available without retraining baseline patient models. Additional domains, such as early postoperative recovery metrics, can be incorporated by extending the stacking layer, positioning the framework as reusable infrastructure rather than a single-purpose model.

\subsection{Limitations}
Several limitations warrant consideration. First, although data were drawn from a statewide health information exchange encompassing over 100 healthcare systems, external validation at independent institutions is required to assess transportability across settings with differing documentation practices, patient populations, and EHR configurations. Temporal validation (AUROC 0.915) provides encouraging evidence of stability across time periods within the same data source, but the calibration slope of 1.330 on the temporal test set indicates that recalibration may be necessary when deploying across different time periods or institutions.

Second, outcome ascertainment relied on structured CAM documentation and ICD coding, potentially under-detecting cases lacking formal assessment; however, the dual-confirmation strategy was designed to maximize specificity.

Third, the case--control design yields artificial 50\% prevalence, which may influence discrimination estimates. To quantify this effect, we evaluated the model under simulated 1:2 and 1:3 case-to-control ratios, observing AUROC of 0.887 and 0.867, respectively, demonstrating graceful degradation rather than dependence on balanced sampling. Additionally, matching on age, sex, race, and surgery year removes these variables' independent discriminative contribution, potentially underestimating the framework's full discrimination in unmatched populations where these known POD risk factors would contribute additional signal. We addressed the calibration implications through prevalence adjustment and weighted decision curve analysis, but prospective validation in an unmatched cohort remains necessary.

Fourth, patients with preexisting dementia were excluded to isolate incident postoperative delirium from chronic cognitive impairment, which has fundamentally different pathophysiology and clinical management implications~\cite{ref19}. This exclusion limits the model's applicability to a critical high-risk population. Future work should develop dedicated models for patients with baseline cognitive impairment, potentially incorporating cognitive assessments as additional predictors.

Fifth, the current framework incorporates preoperative and intraoperative data; extension to postoperative predictors would require additional validation. These limitations pertain primarily to the specific POD application rather than the framework architecture, which accommodates alternative sampling strategies, outcome definitions, and data sources.

\subsection{Future Directions}
Future work should prioritize external validation at independent institutions with attention to calibration stability under distribution shift. Prospective evaluation of whether framework-guided decision support improves clinical outcomes, such as targeted delirium prevention protocols, represents a critical translational step. Application of the framework to additional perioperative outcomes (e.g., acute kidney injury, respiratory failure) would test whether domain contributions shift as hypothesized in Table~\ref{tab:domains}. Integration with real-time EHR systems would enable dynamic risk updating throughout the perioperative trajectory.

\section{Conclusion}

We developed and internally validated a domain-structured ensemble framework for perioperative outcome prediction that organizes predictors into patient-related, surgery-related, and anesthetics-related domains and integrates domain-level predictions through stacked ensemble learning. Applied to postoperative delirium, the framework achieved strong discrimination (AUROC 0.899; 95\% CI: 0.891--0.906), excellent calibration, and positive clinical utility across decision thresholds. Domain ablation confirmed that the multi-domain architecture improves probability estimation beyond a surgery-only model, and temporal validation demonstrated maintained discrimination on held-out post-2017 data (AUROC 0.915). The modular architecture supports transparent risk attribution, outcome-agnostic deployment, and dynamic updating as perioperative data accrue. With external validation, this approach may serve as a scalable foundation for interpretable, calibration-aware perioperative clinical decision support across multiple postoperative outcomes.

\end{document}